\documentclass[11pt]{article}

\usepackage[T1]{fontenc}
\usepackage[utf8]{inputenc}
\usepackage{lmodern}
\usepackage{textcomp}
\IfFileExists{microtype.sty}{\usepackage{microtype}}{}

\usepackage{amsmath,amssymb}
\usepackage[margin=1in]{geometry}
\usepackage{graphicx}
\usepackage{authblk}
\usepackage{caption}
\usepackage{subcaption}
\usepackage{float}
\usepackage{parskip}
\usepackage[unicode]{hyperref}
\PassOptionsToPackage{hyphens}{url}
\hypersetup{
  pdftitle={AI-driven Optimisation of Quality of Recovery (QoR) in Remote Patient Monitoring},
  pdfauthor={L.-H. Lin, Y. Liu, P. Khetrapal, R. Stafford, J. Kelly, I. Drobnjak},
  pdfkeywords={Artificial intelligence, Digital health, Remote patient monitoring, Patient engagement, Quality of Recovery},
  colorlinks=true,
  linkcolor=black,
  citecolor=black,
  urlcolor=blue
}

\title{\bfseries AI-driven Optimisation of Quality of Recovery (QoR)\\ in Remote Patient Monitoring%
\thanks{Accepted as a poster at AI in Medicine 2026 (Polish Institute for Evidence Based Medicine), a non-archival venue; the accepted abstract is available online~\cite{piebm2026}. This is the authors' preprint version.}}

\author[1]{Yansong Liu\textsuperscript{$\dagger$}}
\author[1]{Li-Hsi (Sonny) Lin\textsuperscript{$\dagger$}}
\author[1,2]{Pramit Khetrapal}
\author[1,2]{Ronnie Stafford}
\author[1,2]{John Kelly}
\author[1]{Ivana Drobnjak\textsuperscript{$\ddagger$}}
\affil[1]{University College London, London, United Kingdom}
\affil[2]{Ethera Health Ltd, London, United Kingdom}

\date{\today}

\begin{document}
\maketitle

\renewcommand{\thefootnote}{\arabic{footnote}}
\setcounter{footnote}{0}
\footnotetext{\textsuperscript{$\dagger$}These authors contributed equally to this work.}
\footnotetext{\textsuperscript{$\ddagger$}Correspondence to
	\href{mailto:i.drobnjak@ucl.ac.uk}{i.drobnjak@ucl.ac.uk}.}

\begin{abstract}
	\noindent Remote patient monitoring depends on patient-reported data to capture
	the subjective dimension of recovery that devices cannot measure. The Quality
	of Recovery (QoR-15) survey is the gold-standard instrument for this purpose.
	It was designed and validated for occasional in-hospital assessment, yet
	remote monitoring now administers it to patients daily. In our own
	post-surgical deployment, only 55\% of patients submitted the survey more than 14 days
	of 30 monitoring days~\cite{halox2025}. We developed QoR-compact, a five-item
	daily input for the RPM prediction pathway. Setting a deployment-driven target
	of one-third of the daily items, we exhaustively evaluated all 3{,}003
	five-question subsets of the QoR-15 and tested whether the best of them
	matches the full instrument in predicting near-term postoperative recovery
	severity. QoR-compact achieves a mean AUC-ROC of 0.968 (95\% CI 0.915--0.988),
	statistically comparable to the 0.964 baseline obtained with one-third of the
	items. Patient-level backtesting indicates that it tracks readmission events
	as faithfully as the full form. Its five items span the physical and
	psychological axes of recovery: Q3 (feeling rested), Q9 (feeling comfortable
	and in control), Q10 (general well-being), Q12 (severe pain), and Q14
	(feeling worried or anxious). The QoR-15 remains the gold-standard measure of
	recovery; QoR-compact complements it as a shorter daily input designed for
	prediction. This parity provides the basis for a prospective study of whether
	a lighter daily input is, in turn, completed more consistently. External
	validation on larger cohorts is required before clinical use.
\end{abstract}


\section{Introduction}
Artificial intelligence remote patient monitoring (AI-RPM) facilitates the
ongoing transition towards patient-centred healthcare by extending elements of
hospital care into patients' homes~\cite{shaik2023}. AI-RPM combines wearable
sensors, patient-reported outcome measures (PROMs), and machine-learning models
to enable continuous, data-driven recovery assessment outside the clinical
setting~\cite{liu2025multimodal}. Its promise is substantial: earlier detection
of clinical deterioration, reduced readmission burden, and greater patient
engagement in their own recovery.

Among the PROMs used in postoperative care, the Quality of Recovery (QoR-15) is
a 15-item survey that has become the gold-standard instrument for assessing
postoperative recovery~\cite{stark2013}. Developed and psychometrically
evaluated by Stark and Myles, it spans both the physical and psychological
dimensions of recovery, and a subsequent systematic review and meta-analysis
confirmed its reliability and responsiveness across diverse surgical
populations~\cite{myles2022}. These properties have led to its adoption in
several RPM systems, where it captures the subjective recovery signal that
wearables alone cannot provide.

The QoR-15 was, however, designed for episodic in-hospital use, typically
administered once on the day of or after surgery~\cite{stark2013}. Remote
monitoring inverts this assumption: patients are asked to complete the full
15-item survey daily, often for weeks after discharge. This daily repetition
imposes a substantial response burden that erodes compliance over
time: feasibility studies of app-based home QoR-15 monitoring have reported
such attrition~\cite{semple2015}, and randomised trials of smartphone-based
postoperative monitoring have likewise identified engagement as the limiting
factor~\cite{temple2023}. In our own post-surgical deployment, only 55\% of
patients submitted the survey more than 14 days of 30 monitoring days~\cite{halox2025}. The
resulting missingness degrades the predictive models that depend on it. To our
knowledge, no study has optimised the QoR-15 specifically for the demands of
AI-RPM.

We therefore investigated whether a shorter subset of the QoR-15, termed
QoR-compact, can be identified that retains the predictive utility of the full
instrument for near-term postoperative recovery severity. A two-thirds reduction
in daily items, if achievable without predictive loss, would reduce the per-day
response burden (a necessary, though not sufficient, step) towards improving
AI-RPM data quality.

To answer this question, we drew on the prospective HALO-Surgery cohort
(IRAS 284073), in which patients submitted daily QoR-15 surveys through a
remote-monitoring platform following abdominal or thoracic cancer
surgery~\cite{halox2025}. Under a pre-specified five-item budget (one-third of
the original instrument), we exhaustively evaluated all
$\binom{15}{5}=3{,}003$ five-question subsets alongside the full 15-item
baseline, using XGBoost multiclass classifiers trained to predict ordinal
recovery classes from 14-day sliding windows. The final QoR-compact was derived
from the items most consistently represented among the top-performing subsets,
and was backtested at the patient level against independent hospital
readmission events.

\section{Materials and Methods}
\label{sec:methods}
We derived QoR-compact from the HALO-Surgery monitoring data and validated it
against patient-level recovery outcomes.

\begin{figure}[htbp]
	\centering
	\includegraphics[width=\linewidth]{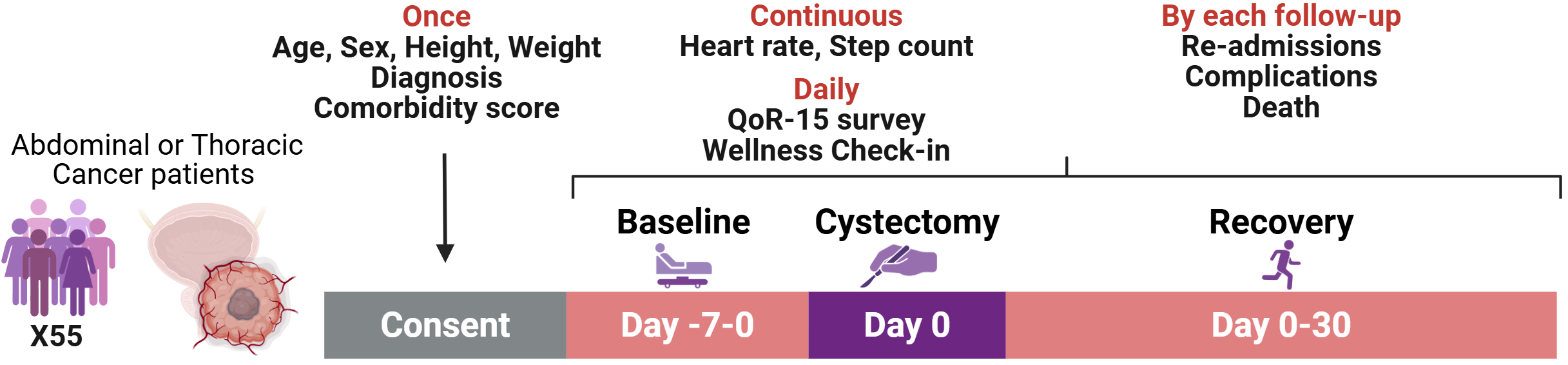}
	\caption{\textbf{Clinical data-collection workflow of the HALO-Surgery
			study.} Eligible patients undergoing abdominal or thoracic cancer surgery
		were enrolled, discharged with a remote-monitoring device, and asked to
		complete the QoR-15 survey daily; the submissions were streamed to the
		HALO platform for analysis.}
	\label{fig:halo-flow}
\end{figure}

\paragraph{Data collection.}
QoR-15 records were drawn from the prospective HALO-Surgery study (IRAS 284073).
Patients discharged after abdominal or thoracic cancer surgery completed the
survey daily through a remote-monitoring platform
(Figure~\ref{fig:halo-flow}). This is the same deployment whose completion
pattern motivated this work, as reported in the Introduction.

\begin{figure}[htbp]
	\centering
	\includegraphics[width=\linewidth]{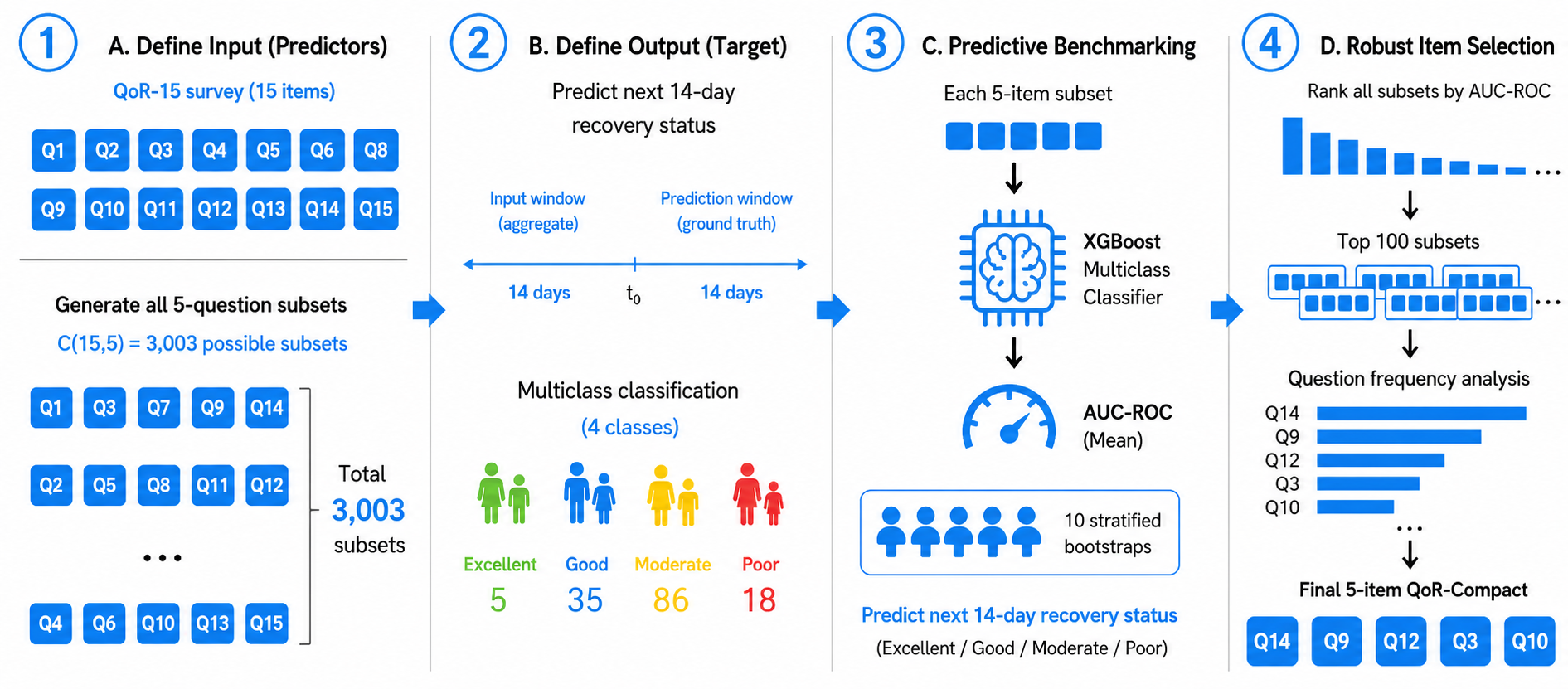}
	\caption{\textbf{Overview of the analysis.} (1) The 15 QoR-15 items
		generate all $\binom{15}{5}=3{,}003$ five-question input subsets.
		(2) The prediction target is the patient's recovery class over the
		14 days following the input window, mapped to four ordinal categories.
		(3) Each subset is benchmarked with an XGBoost multiclass classifier
		under 10 stratified bootstrap resamples. (4) Subsets are ranked by
		AUC-ROC; the five items most consistently appearing among the top-100
		subsets constitute QoR-compact.}
	\label{fig:pipeline}
\end{figure}

\paragraph{Preprocessing.}
Each patient's longitudinal record was processed with a sliding window that
advances one day at a time. At each position, the preceding 14 days form an
\emph{input window}. The feature vector is the mean score of each question over
the input window (Figure~\ref{fig:pipeline}, Panel~1), preserving the
instrument's native 0--10 scale while summarising the patient's recent recovery
trajectory.

\paragraph{Exploratory collinearity analysis.}
Before any modelling, we ran a pre-specified collinearity analysis across the
15 items to assess whether a smaller, well-chosen subset could plausibly retain
the predictive signal of the full survey. Running this analysis before training
keeps the motivation for compaction independent of the modelling outcome.
Pairwise Spearman rank correlations $\rho$ and variance-inflation factors were
computed across all patient-day observations. The Spearman correlations were
summarised as a lower-triangle heatmap
(Figure~\ref{fig:corr-matrix}, Supplementary Material) and converted into a
hierarchical clustering dendrogram using the distance $d=1-|\rho|$ with average
linkage, so that redundant items group into visible branches
(Figure~\ref{fig:collinearity}). Findings are reported in Results.

\paragraph{Label construction.}
The prediction target is the patient's near-future recovery state
(Figure~\ref{fig:pipeline}, Panel~2). The 14 days
following each input window form an \emph{output window}. The aggregate QoR
score over the output window (range 0--150) is mapped to one of four ordinal
recovery classes using clinically established QoR-15
thresholds~\cite{kleif2018}: Excellent, Good, Moderate, and Poor. Each valid
window position yields one input--output pair. Cohort sizes and class counts
are reported in Results.

\paragraph{Exhaustive evaluation and selection.}
Under the five-question budget fixed above, every one of the
$\binom{15}{5}=3{,}003$ possible 5-question subsets was evaluated, together with
the full 15-question baseline. For each subset, an XGBoost multiclass classifier
was trained to predict the four-class recovery label
(Figure~\ref{fig:pipeline}, Panel~3). All models were trained
and evaluated under identical conditions across 10 stratified bootstrap
resamples, so that any performance difference is attributable solely to question
choice. Models were ranked by one-vs-rest weighted Area Under the Receiver
Operating Characteristic curve (AUC-ROC), with 95\% confidence intervals taken
from the bootstrap percentiles. We did not adopt the single best-ranked subset,
which may reflect idiosyncrasies of one dataset. Instead, we identified
questions by their consistency across the top-100 subsets
(Figure~\ref{fig:pipeline}, Panel~4). Each subset draws 5
of 15 questions, so any question would appear $5\times100/15\approx33.3$ times
by chance. Within the five-question budget, the five questions most consistently
over-represented among the top-100 subsets were taken to constitute the final
QoR-compact.

\paragraph{Analysis and visualisation.}
All analyses were performed in Python with scikit-learn~\cite{scikit-learn2011}
for metric computation and data splitting. For the exhaustive evaluation
(Figure~\ref{fig:subsets}), each model's bootstrap mean AUC-ROC is shown as a
horizontal bar, with asymmetric 95\% CI error bars taken from the empirical
2.5th and 97.5th percentiles of the per-split AUC values; this preserves the
skewed bootstrap distribution, unlike a $\pm$ standard error. The baseline
15-item model is shown as a separate reference bar. The frequency with which
each item appears among the top-100 subsets is plotted as a horizontal bar chart
(Figure~\ref{fig:frequency}), against a by-chance expectation of
$5\times100/15\approx33.3$ occurrences. For qualitative patient-level validation
(Figure~\ref{fig:backtesting}), per-patient dual-axis trajectories plot the
daily full QoR-15 total (0--150, left axis) against the QoR-compact total
(0--50, right axis) over postoperative days 0--30, with hospital readmission and
complication events superimposed as vertical markers.

\newpage
\section{Results}
After cleaning, 1{,}035 daily QoR-15 submissions from the eligible cohort were
retained for analysis. Applying the sliding-window procedure of
Section~\ref{sec:methods} yielded 144 input--output pairs. These fell into the
four recovery classes as 5 Excellent, 35 Good, 86 Moderate, and 18 Poor.

\begin{figure}[htbp]
	\centering
	\includegraphics[width=0.9\linewidth]{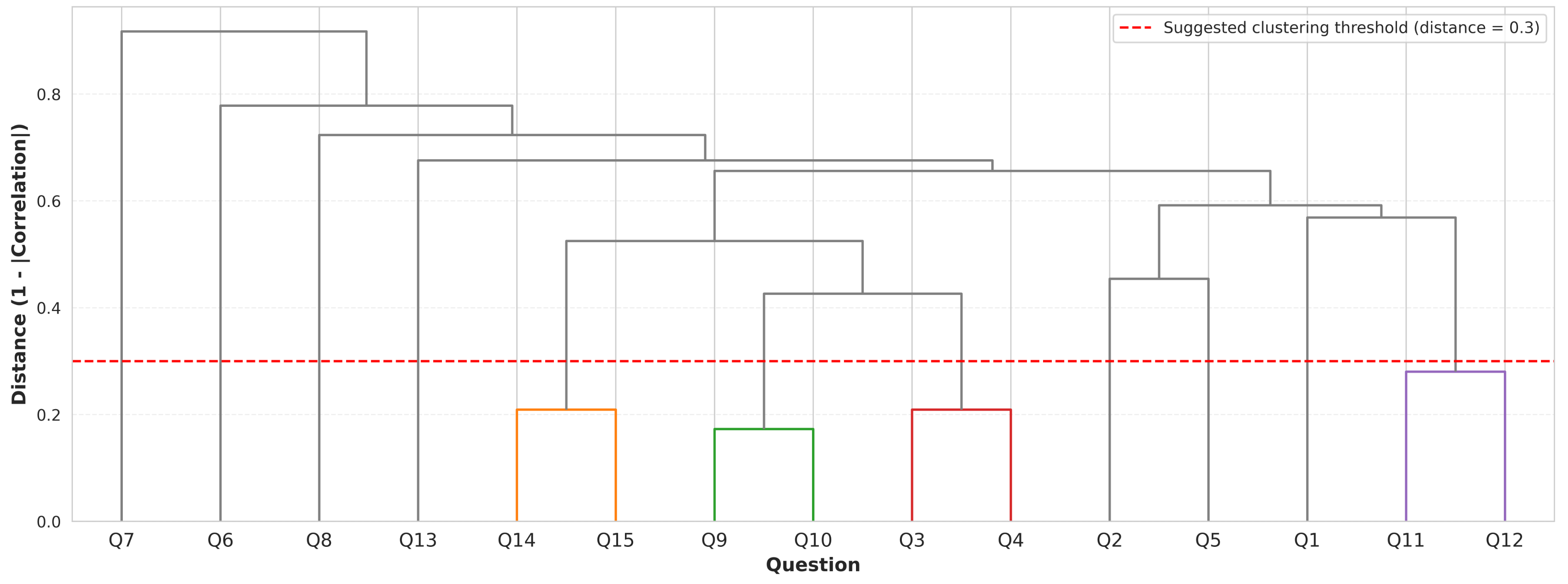}
	\caption{\textbf{Hierarchical clustering of the 15 QoR-15 items.} The dendrogram
		uses distance $=1-|\rho|$ with average linkage on the pairwise Spearman
		correlations. Tight item clusters pair questions that probe the same
		domain: Q3 (feeling rested) with Q4 (good sleep), Q11 (moderate pain)
		with Q12 (severe pain), Q9 (feeling comfortable and in control) with
		Q10 (general well-being), and Q14 (feeling worried or anxious) with
		Q15 (feeling sad or depressed). This structured redundancy motivates
		the search for a compact form. The underlying correlation matrix is
		shown in Figure~\ref{fig:corr-matrix} (Supplementary Material).}
	\label{fig:collinearity}
\end{figure}

Before searching for a shorter form, we asked whether the 15 items of the QoR-15
carry independent information. Pairwise Spearman correlations and
variance-inflation factors revealed several tight item clusters whose
information was largely recoverable from neighbouring items
(Figure~\ref{fig:collinearity}). The tightest pairs, such as Q9--Q10, Q3--Q4,
and Q14--Q15, group semantically related questions, indicating that the
redundancy is structured rather than random.

\begin{figure}[H]
	\centering
	\begin{subfigure}[t]{0.48\linewidth}
		\centering
		\includegraphics[width=\linewidth]{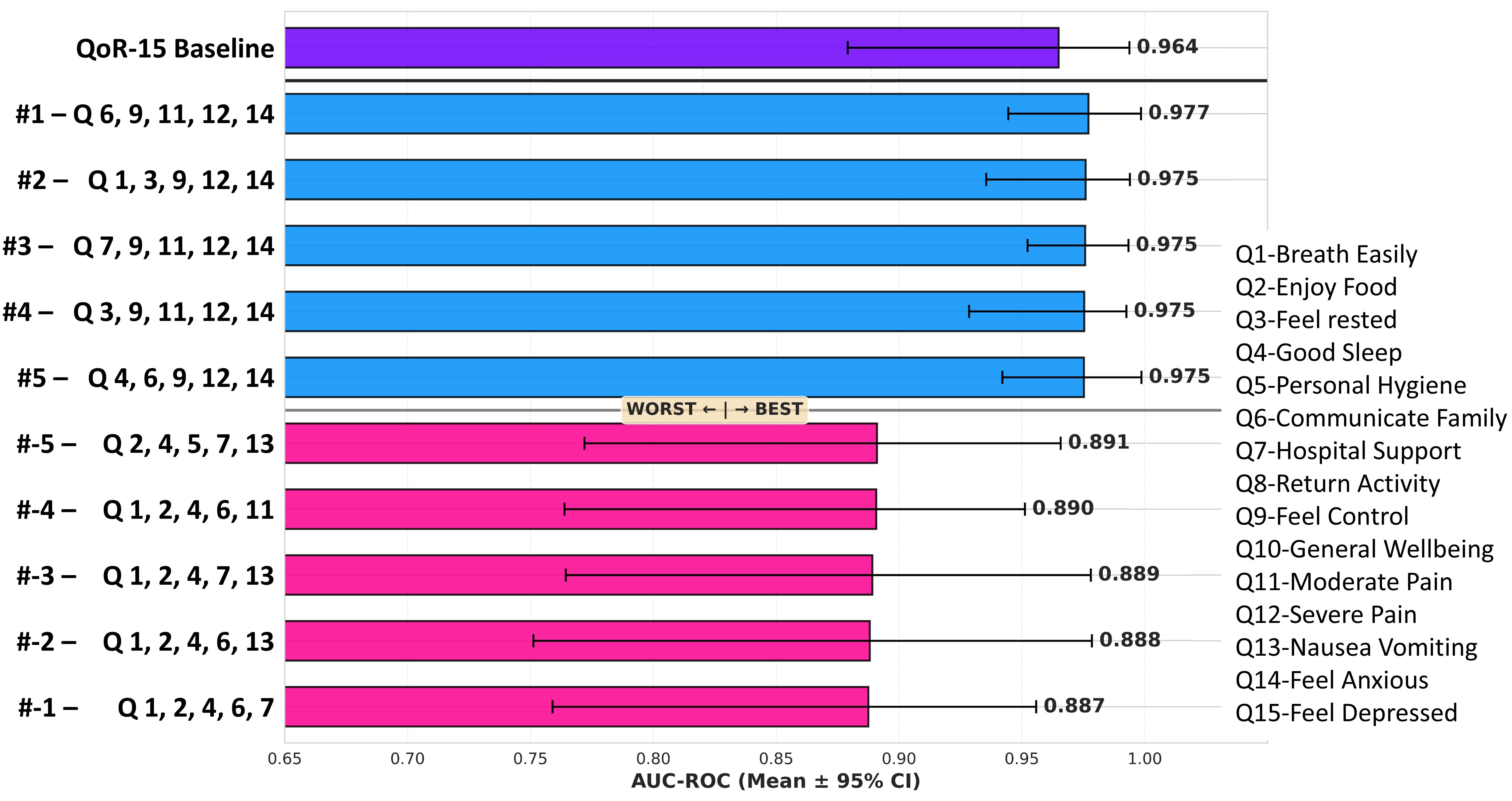}
		\caption{\textbf{Best and worst 5-question subsets.} Mean
			AUC-ROC with 95\% confidence intervals. The full 15-item
			baseline (purple) is 0.964; the top-ranked subsets (blue)
			reached 0.977; the worst (red) fell well below.}
		\label{fig:subsets}
	\end{subfigure}\hfill
	\begin{subfigure}[t]{0.48\linewidth}
		\centering
		\includegraphics[width=\linewidth]{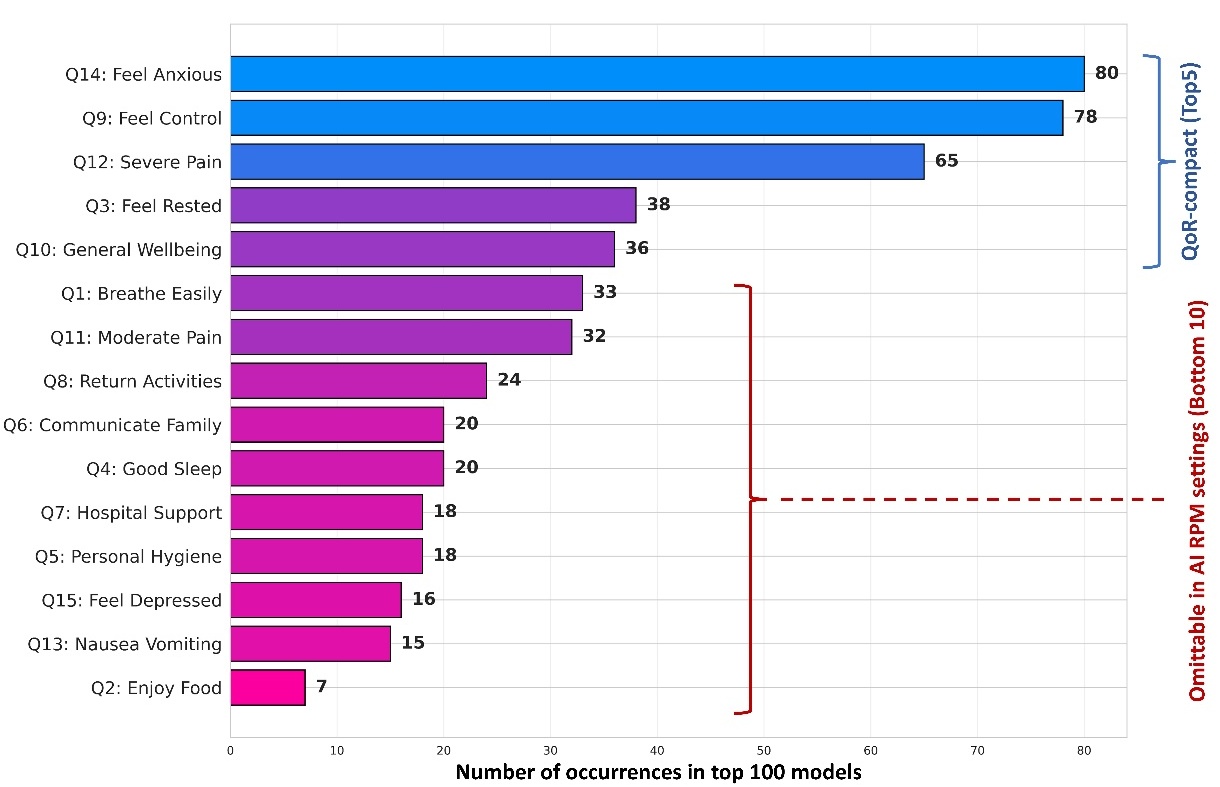}
		\caption{\textbf{Item frequency among the top-100 subsets.}
			The five most frequent items---Q14 (feeling worried or
			anxious), Q9 (feeling comfortable and in control), Q12
			(severe pain), Q3 (feeling rested), and Q10 (general
			well-being)---constitute QoR-compact.}
		\label{fig:frequency}
	\end{subfigure}
	\caption{Exhaustive subset evaluation and robust item selection.}
\end{figure}

The full 15-item QoR-15 set a strong baseline (Figure \ref{fig:subsets} purple bar), reaching a mean AUC-ROC of 0.964
(95\% CI 0.879--0.994) for predicting near-term recovery severity from the
aggregated 14-day window. The exhaustive search then showed that this predictive
signal is not evenly distributed across the 15 items. The best 5-question
subsets reached 0.977 (95\% CI 0.944--0.998. Figure \ref{fig:subsets} blue bars), above the full-form ceiling,
whereas the worst fell well below it (Figure \ref{fig:subsets} red bars). The approximately 0.09 AUC-ROC gap between
best and worst tracked which items the subsets shared
(Figure~\ref{fig:subsets}). Subsets built on less informative items such as Q2
(enjoy food) and Q4 (good sleep) ranked consistently lower.

To move from the existence of strong subsets to a single deployable short form,
we ranked the 3{,}003 subsets and counted how often each question appeared among
the top 100 (Figure~\ref{fig:frequency}). Three questions---Q14 (feeling worried
or anxious), Q9 (feeling comfortable and in control), and Q12 (severe
pain)---appeared well above the by-chance threshold of $\approx 33.3$. The next
two, Q3 (feeling rested) and Q10 (general well-being), appeared only slightly
above it. Under the pre-specified five-question budget, these five
highest-frequency questions constitute QoR-compact. Evaluated as a fixed
instrument, it achieved a mean AUC-ROC of 0.968 (95\% CI 0.915--0.988), with a
confidence interval that overlaps substantially with the 0.964 baseline.

\begin{figure}[H]
	\centering
	\includegraphics[width=0.9\linewidth]{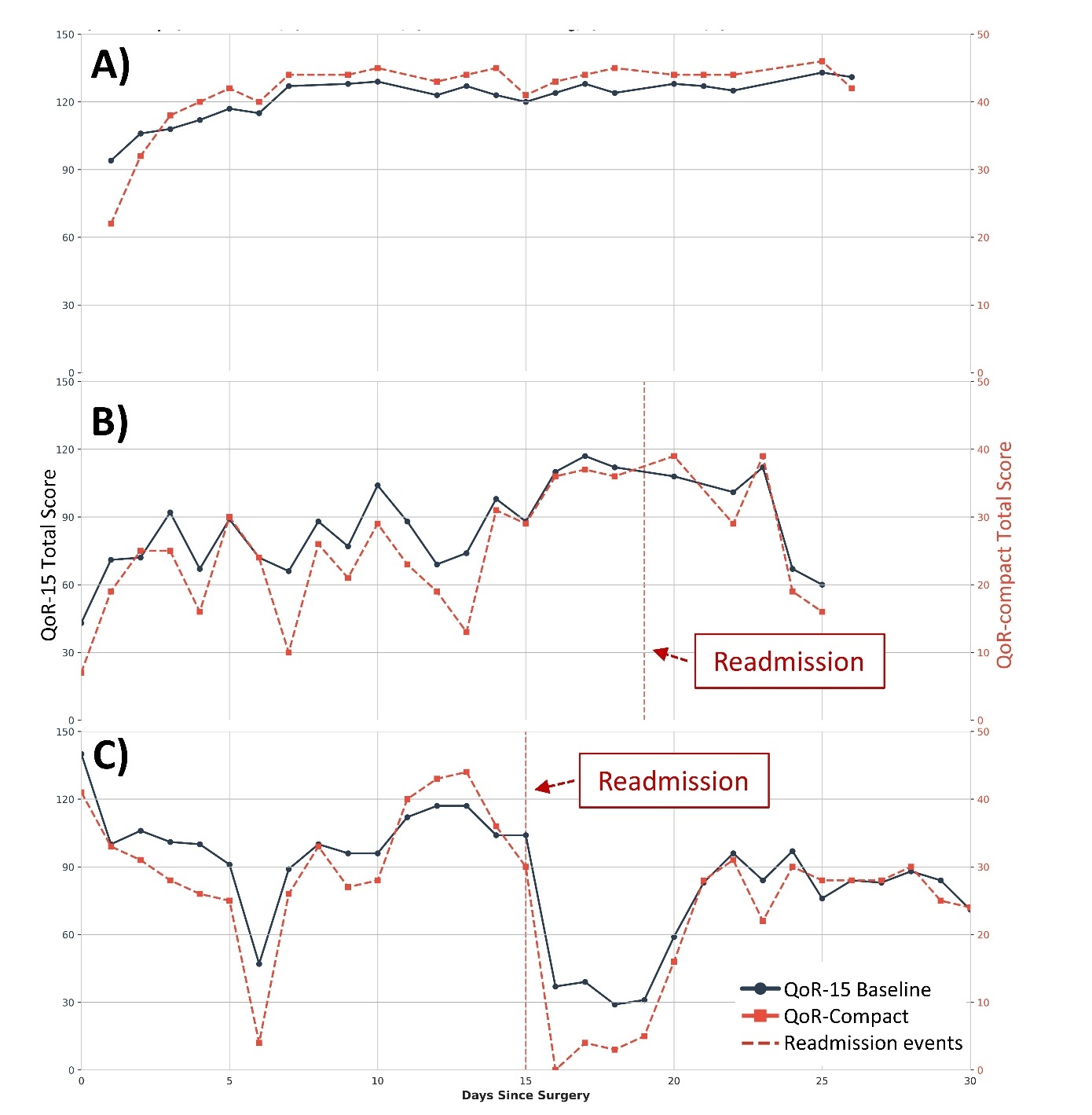}
	\caption{\textbf{Longitudinal backtesting of QoR-15 baseline versus QoR-compact
			scores.} Postoperative recovery trajectories over 30 days are shown for three
		individual patients (Panels A, B, and C). The full QoR-15 total score (solid
		line, left axis; max 150) is plotted alongside the derived 5-question
		QoR-compact total score (dashed line, right axis; max 50). The QoR-compact
		demonstrates strong alignment with the baseline survey's daily variance. Panel
		A depicts an uncomplicated, steady recovery. In Panels B and C, vertical dashed
		lines denote hospital readmissions, which are accurately captured by concurrent
		sharp declines and volatility in both the baseline and compact scores.}
	\label{fig:backtesting}
\end{figure}

To test whether this parity holds against an external clinical anchor, we
backtested the two instruments day by day over the first 30 postoperative days
(Figure~\ref{fig:backtesting}). In an uncomplicated recovery (Panel A), the
QoR-compact trajectory closely tracks the full QoR-15. For the two patients who
were readmitted (Panels B and C), both instruments register concurrent declines
at the readmission event, and QoR-compact is slightly more reactive at points of
acute deterioration.

\section{Discussion and Limitations}
Using one-third of the QoR-15 items, QoR-compact reached predictive performance comparable
with the full 15-item instrument. Its
point estimate was marginally higher (0.968 vs.\ 0.964), with narrower
bootstrap intervals.

This performance is consistent with the pre-specified collinearity analysis.
The 15 QoR-15 items form tight semantic clusters (pain, rest, affect,
wellbeing) whose information is largely redundant, so pruning within clusters
need not sacrifice signal. The exhaustive search confirmed that the
remaining signal is not evenly distributed: whether a five-item form matches
the full instrument depends on which items are retained. This distinguishes the present predictive-compaction approach from
traditional psychometric short-form development~\cite{coste1997}, which selects items to
preserve latent construct coverage rather than predictive performance against
an external criterion.

The selected items are clinically coherent. Q12 (severe pain) and Q3 (feeling
rested) track the physical trajectory of recovery; uncontrolled pain and
disrupted rest are well-established antecedents of postoperative
deterioration~\cite{kehlet2002}. Q14 (anxiety), Q9 (sense of control), and Q10 (general
wellbeing) capture the psychological dimension. Post-surgical anxiety and a
perceived loss of control are recognised correlates of poor recovery
outcomes~\cite{myles2022}. These are also the items the backtests show to be
most reactive at points of acute deterioration, consistent with their role in
an early-warning context.

The ten items excluded from QoR-compact also merit attention. Several,
including Q1 (breathing), Q5 (normal activities), and Q8 (self-care), carry
less
discriminative signal because they recover towards ceiling early in the
postoperative course and therefore vary little across the 14-day prediction
window. Others, such as Q13 (nausea), are episodic and poorly summarised by a
windowed mean. Q7 (support from hospital staff) reflects healthcare delivery
rather than the patient's own recovery state. The exhaustive search confirmed
that these items contribute less marginal predictive information in this
cohort.

These findings have implications for how PROMs serve remote monitoring.
Earlier studies of app-based postoperative monitoring administered the full
QoR-15 at home without specifically evaluating whether all 15 items are
necessary for the predictive tasks that monitoring ultimately
serves~\cite{semple2015, temple2023}. The present results suggest that, for
prediction rather than measurement, a substantially shorter input can suffice.
The QoR-15 remains the appropriate instrument for measuring
recovery~\cite{myles2022};
QoR-compact is designed for AI RPM applications. The two are complementary: the full
form could anchor periodic clinical assessment, while the short form serves as
the daily model input, reducing the response burden that drives the
non-compliance observed in our own deployment~\cite{halox2025}.

Several boundaries of this work merit attention. The shared derivation of
features and labels from the QoR-15 could in theory inflate the AUC-ROC
through instrument-internal redundancy. That said, the backtests against
hospital readmission, an endpoint entirely independent of the
survey, confirm that both instruments track genuine clinical deterioration.
The cohort is single-centre and modest. Crucially, this is only an exploratory study on a concrete clinical question that has not been previously addressed. The contribution and implications of this work should be interpreted beyond the immediate clinical context. Finally, the selected items were optimised for
this cohort's case mix. The exhaustive-search framework is, by design,
portable: it can be re-deployed on any new cohort to produce a refreshed
ranking.

\section{Conclusion and Future Work}
Future work will test QoR-compact on larger, multi-centre prospective cohorts
spanning additional cancer types and surgical modalities, with a formal
assessment against the full QoR-15. It will explore sequence
models that operate on raw daily trajectories and integrate survey responses
with continuous wearable-derived signals within a broader multimodal
remote-monitoring framework~\cite{liu2025multimodal}. A prospective engagement
study is needed to determine whether the two-thirds reduction in daily items
translates into measurably higher compliance; the parity reported here removes
the predictive objection to asking that question. In summary, a five-item
subset of the QoR-15 can match the full instrument's near-term predictive
performance, providing a practical basis for lighter daily inputs in AI-RPM.
\section*{Disclosures}
\addcontentsline{toc}{section}{Disclosures}
Pramit Khetrapal, Ronnie Stafford, and John Kelly are co-founders and/or
directors of Ethera Health Ltd, UK, which developed the remote patient monitoring
platform used in this work. All
study design, data collection, analysis, and interpretation were conducted
independently of company management.


\appendix
\renewcommand{\thefigure}{S\arabic{figure}}
\setcounter{figure}{0}
\section*{Supplementary Material}
\addcontentsline{toc}{section}{Supplementary Material}

\begin{figure}[htbp]
	\centering
	\includegraphics[width=0.7\linewidth]{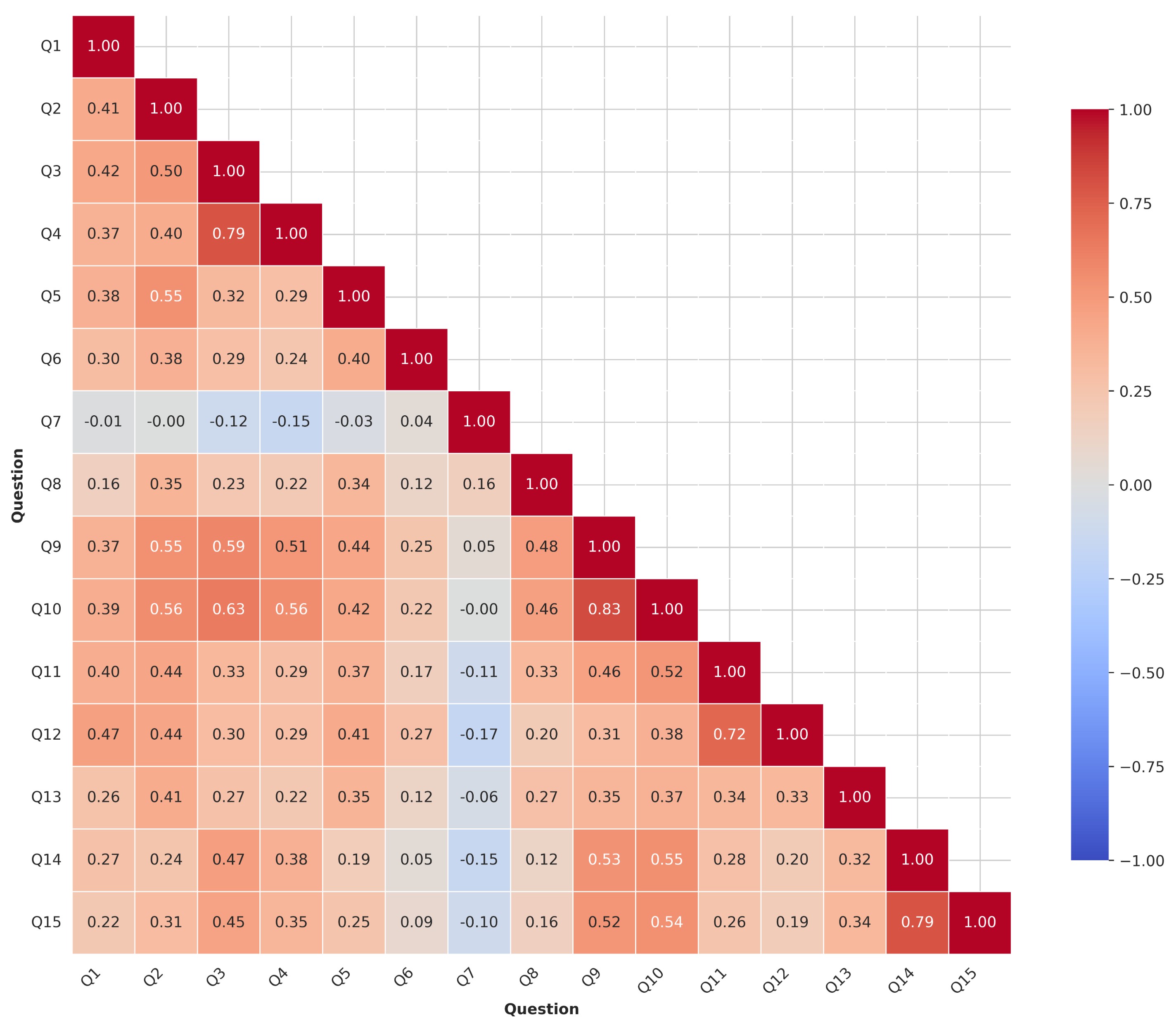}
	\caption{\textbf{Lower-triangle Spearman correlation matrix of the 15 QoR-15
			items} in the HALO-Surgery cohort. Pairwise rank correlations $\rho$ were
		computed across all patient-day observations. Strong positive correlations
		(dark red) between items such as Q9--Q10, Q3--Q4, Q14--Q15, and Q11--Q12
		reflect the redundancy that the dendrogram in
		Figure~\ref{fig:collinearity} summarises as tight clusters.}
	\label{fig:corr-matrix}
\end{figure}


\begin{thebibliography}{9}

	\bibitem{shaik2023}
	Shaik T, Tao X, Higgins N, et al.
	Remote patient monitoring using artificial intelligence: Current state,
	applications, and challenges.
	\emph{WIREs Data Mining and Knowledge Discovery}. 2023;13(2):e1485.
	doi:\url{https://doi.org/10.1002/widm.1485}

	\bibitem{churruca2021}
	Churruca K, Pomare C, Ellis LA, et al.
	Patient-reported outcome measures (PROMs): A review of generic and
	condition-specific measures and a discussion of trends and issues.
	\emph{Health Expectations}. 2021;24(4):1015--1024.
	doi:\url{https://doi.org/10.1111/hex.13254}

	\bibitem{stark2013}
	Stark PA, Myles PS, Burke JA.
	Development and psychometric evaluation of a postoperative quality of recovery
	score: the QoR-15.
	\emph{Anesthesiology}. 2013;118(6):1332--1340.
	doi:\url{https://doi.org/10.1097/ALN.0b013e318289b84b}

	\bibitem{myles2022}
	Myles PS, Shulman MA, Reilly J, Kasza J, Romero L.
	Measurement of quality of recovery after surgery using the 15-item quality of
	recovery scale: a systematic review and meta-analysis.
	\emph{British Journal of Anaesthesia}. 2022;128(6):1029--1039.
	doi:\url{https://doi.org/10.1016/j.bja.2022.03.009}

	\bibitem{semple2015}
	Semple C, Bhatt B, Sharpe S, et al.
	Using a mobile app for monitoring post-operative quality of recovery of
	patients at home: a feasibility study.
	\emph{Journal of Medical Internet Research}. 2015;17(7):e168.
	doi:\url{https://doi.org/10.2196/jmir.3851}

	\bibitem{temple2023}
	Temple-Oberle C, Shea-Budgell MA, Tan M, et al.
	Effect of Smartphone App Postoperative Home Monitoring After Oncologic
	Surgery on Quality of Recovery: A Randomized Clinical Trial.
	\emph{JAMA Surgery}. 2023;158(11):1181--1188.
	doi:\url{https://doi.org/10.1001/jamasurg.2023.4145}

	\bibitem{kleif2018}
	Kleif J, G\"ogenur I.
	Severity classification of the quality of recovery-15 score---An observational
	study.
	\emph{Journal of Surgical Research}. 2018;225:101--107.
	doi:\url{https://doi.org/10.1016/j.jss.2017.12.040}

	\bibitem{liu2025multimodal}
	Liu Y, Stafford R, Khetrapal P, et al.
	Multi-Modal AI for Remote Patient Monitoring in Cancer Care.
	\emph{arXiv preprint} arXiv:2512.00949. 2025.
	\url{https://arxiv.org/abs/2512.00949}

	\bibitem{halox2025}
	Stafford R, Liu Y, Khetrapal P, Carvalho G, Kocadag H, Surrao D, McBain H,
	Winter P, Jackson-Spence F, Powles T, Kelly JD, Drobnjak I.
	HALO-X: A full-stack remote patient monitoring platform for post-cancer
	recovery. Manuscript under review, 2025.

	\bibitem{scikit-learn2011}
	Pedregosa F, Varoquaux G, Gramfort A, et al.
	Scikit-learn: Machine Learning in Python.
	\emph{Journal of Machine Learning Research}. 2011;12:2825--2830.

	\bibitem{coste1997}
	Coste J, Guillemin F, Pouchot J, Fermanian J.
	Methodological approaches to shortening composite measurement scales.
	\emph{Journal of Clinical Epidemiology}. 1997;50(3):247--252.
	doi:\url{https://doi.org/10.1016/S0895-4356(97)90533-8}

	\bibitem{kehlet2002}
	Kehlet H, Wilmore DW.
	Multimodal strategies to improve surgical outcome.
	\emph{American Journal of Surgery}. 2002;183(6):630--641.
	doi:\url{https://doi.org/10.1016/S0002-9610(02)00412-3}

	\bibitem{piebm2026}
	Lin LH, Liu Y, Khetrapal P, et al.
	AI-driven Optimisation of Quality of Recovery (QoR) in Remote Patient
	Monitoring [abstract]. Accepted as a poster at AI in Medicine 2026, Polish
	Institute for Evidence Based Medicine. Available at:
	\href{https://piebm.org/abstracts/ai-driven-optimisation-of-quality-of-recovery-qor-in-remote-patient-monitoring/}{piebm.org/abstracts/qor-remote-patient-monitoring}

\end{thebibliography}
\end{document}